# LEXICAL BUNDLES IN COMPUTATIONAL LINGUISTICS' ACADEMIC LITERATURE

Adel Rahimi

adelr@adelr.ir

## 1. Introduction

Corpus linguistics has been in the spotlight for the last decade, with the usage of modern computers and technologies deeper understanding of languages can be obtained. Corpus linguistics helps language teaching for acquiring a better view for language. Language teachers will know what sequence of words and patters tend to co-occur. Unlike previous enormous lists of words which students were forced to memorize them, and were seldom used and typically were forgotten in a short period of time lexical bundles can help language teachers to teach more effectively, and learners can be more fluent in the second language. The first studies in lexical bundles include Firth 1964 (Firth, J. R. (1964).) Biber (2004), (Cortes, V. (2004). Native speakers use a formulaic pattern of speech which they are unaware of but learners from other languages use bundles that are affected (transferred) bye their mother tongue and this solely can be problematic and easily recognizable. moreover in academia, when speakers of source language trying to write, publish, and produce academic literature in the target language their lack the fluency and native like features of a native speaker of that language. Lexical bundles (or as called N-grams) are crucial in getting a fluent academic text. In this paper lexical bundles of 1 to 5 tokens from an 8 million word corpus of academic literature from the Computational Linguistics field and its sub topics such as: Speech recognition, Natural Language Processing, Machine Learning, and Information Retrieval have been extracted and analyzed. On the top of that most of typical criteria for exclusion has been applied to the list as well as calculating MI factor for each result to confirm the results and reaching the target bundles for Computational Linguistics.

## 2. Methodology and Data

The corpus of Computational linguistics' academic literature is an 8 million word corpus of Journal publications, books, and theses. These include interdisciplinary topics such as Speech Recognition, Experimental Phonology, Language Models, Machine Learning, Semantics, Syntactic Theory, and Information Retrieval.

Table 2.1 shows the distribution of sources:

| Source | Number of words | Percentage |
|---|---|---|
| Books | ~2million | 25% |
| Journal Articles | ~6million | 75% |

All the books, articles, and these were in PDF format, first combined then edited and all the names of Authors, references, formulas, and tables were removed in order to minimize the "Noise" from the results.

The corpus was saved in txt file and analyzed by KfNgram program for linguistic research. Fletcher, 2007 (Fletcher, W. (2007).

## 2.1 Extracting Lexical Bundles

As biber et al. (1999) states "lexical bundles are sequences of words that commonly go together in natural discourse".

## 2.2 filtering criteria

In order for our data to be in a consistent and logical form and not to be too long and out of the topic or even unusable we have to filter our results. Excluding certain lexical bundles does not mean that they are not lexical bundles or they don't meet the requirements, undoubtedly and technically they are, however it is only for pedagogical purposes these lexical bundles do not have the characteristics to meet the scope and goal of this study or as Salazar D. (2011) says "…additional step was found to be necessary for the study to achieve its primary objective of creating a list of only the pedagogical useful bundles in scientific writing." this exclusion includes bundles that are detached (an integrated model, the operator), bundles that include numeric (House et al 2001).

Table 2.2 adapted from Salazar D. (2011) shows the criteria for exclusion of lexical bundles

| Lexical bundle | Example |
|---|---|
| Fragments of other bundles | there is no |
| Bundles with random number | ## page 12 |
| Meaningless bundles | # and # |
| Noise | pp ###â ### |

Fragments of other bundles: lexical bundles with lower number of chunks can be part of bigger bundles and therefore cannot be used in our research

*can be* is part of a bigger chunk *can be used*

Bundles ending in articles: As can be seen most of the lexical bundles ending in articles (a, an, the) are part of other larger lexical bundles and therefore they are moreless duplicates of the latter e.g. in addition to the, argue in favor of the, the output of the.

bundles with random numbers: bundles including random numbers such as: from gate 2 in 1999

Meaningless bundles: these bundles do not have any meaning and therefore cannot be used in our research examples are: de bot t van els, englewood cliffs nj prentice-hall

Noise: Noise refers to any numeric, symbol, alphabet which neither has meaning, nor is grammatically correct. e.g. â â â â, t 1 t 1

## 3.0 Structure, function and frequency of bundles

This chapter analyses and categorizes lexical bundles by their frequency, function and structure.

## 3.1 Frequency of Lexical Bundles

A total number of 591 bundles across the 8 million corpus of Computational Linguistics discipline have has been extracted. These n-grams have been elected from 1-grams to 5-grams varying in size.

Tables 3.1 shows the distribution of 2-5 part lexical bundles by their size.

| Number of words | Frequency |
|---|---|
| 2 words | 212 |
| 3 words | 211 |
| 4 words | 85 |
| 5 words | 78 |
| | 586 |

These 586 bundles which make up only 1797 tokens consist only 0.02% of the total 8 million corpus; however only the top 5 results from each n-gram category (1 to 5 parts that is a total of only 25 bundles) have occurred more than 126800 times which is approximately 1.6% of the corpus.

Table 3.2 shows the top 50 unfiltered lexical bundles

| | |
|---|---|
| can be | in which |
| to be | use of |
| it is | will be |
| in this | may be |
| for example | the number of |
| number of | in order to |
| such as | in terms of |
| the same | as well as |
| this is | a set of |
| set of | the use of |
| is not | a number of |
| based on | in proceedings of |
| of this | the fact that |
| there is | on the other hand |
| we have | in the case of |
| that is | can be used to |
| of these | on the basis of |
| is to | at the same time |
| there are | in the context of |
| the following | it is possible to |
| going to | at the end of |
| we can | with respect to the |
| which is | in this section we |
| is that | it is important to |
| has been | in the form of |

3.2 Structure of lexical bundles

As biber (1999) observed lexical bundles tend to fall into several structural categories that are related to each other.

Table 3.3 shows different structure classification for lexical bundles adapted from Salazar D. (2011)

| STRUCTURE | EXAMPLES |
|---|---|
| Noun phrase with of-phrase fragment | the end of the, the beginning of the, the base of the, the point of view of |
| Noun phrase with other post-modifier fragments | the way in which, the relationship between the, such a way as to |
| Prepositional phrase with embedded of-phrase fragment | about the nature of as a function of as a result of the, from the point of view of |
| Other prepositional phrase (fragment) | as in the case, at the same time as, in such a way as to |
| Anticipatory it + verb phrase/adjective phrase | it is possible to, it may be necessary to, it can be seen, it should be noted that, it is interesting to note that |
| Passive verb + prepositional phrase fragment | is shown in figure/fig., is based on the, is to be found in |
| Copula be + noun phrase/adjective phrase | is one of the, may be due to, is one of the most |
| (Verb phrase +) that-clause fragment | has been shown that, that there is a, studies have shown that |
| (Verb/adjective +) to-clause fragment | are likely to be, has been shown to, to be able to |
| Adverbial clause fragment | as shown in figure/fig., as we have seen |
| Pronoun/noun phrase + be (+...) | this is not the, there was no significant, this did not mean that, this is not to say |
| Other expressions | that as well as the, may or may not, the presence or absence |

All the lexical bundles extracted from the Computational Linguistics' Academic Literature corpus have been analyzed and put in categories.

### 3.3 function of lexical bundles

Lexical bundles differ in function and therefore as Hyland 2008 categorized them into two functions and 7 sub categories. These sub categories include: Location, Procedure, Quantification, Description, Topic, Transitioning, Resulting, structuring, framing, stance and engagement.

Table 3.6 shows lexical bundles in different functional categories based on Hyland, 2008

| | | |
|---|---|---|
| Research oriented | Location | is shown in figure, as shown in figure, as shown in table, as we will see, are shown in table, |
| | Procedure | in the process of, can be used in, it is difficult to, the same way as, to be used in, |
| | Quantification | a very large number of |
| | Description | , is referred to as, for the purposes of, the degree to which, for the purpose of, to be used in, |
| | topic | shown in the right part, that it is possible to, the value of the, |
| Text oriented | Transition | and on the other hand, on the one hand and, on the one hand |
| | Resultative | more likely to be, is likely to be, are more likely to, can be interpreted as, |
| | Structuring | in terms of their, of the fact that, from the fact that, on the use of, should be noted that, in a variety of, |
| | Framing | the extent to which, it should be noted, the degree to which, in a way that, should be noted that, as a set of, |
| Participant oriented | Stance | is an example of, that can be used, this is going to |
| | engagement | so i'm going to, can be seen in, we would like to, so we're going to, and we're going to, |

Research oriented bundles mostly help writers to structure their ideas and findings. This category consists of several sub-categories that can be explained as follow:

Location: location bundles indicate to a particular time and place.

And it's only ever seen at the end of a sequence

In this section we focus on some of the

Contain discourse information and at the same time offer repetitions of syllables

Procedure: Procedure lexical bundles show how a something is done

Quantification: these bundles talk about numbers and chunks

Description: these bundles try to explain

Topic: these bundles refer to specific topics

Text oriented Lexical Bundles help organizing text and it's message these bundles include:

Transitioning: these bundles show a transiton in the text

Resulting: these bundles show a resultative sentece

structuring: these bundles are for structuring the text

framing: the bundles are for framing the writing

Participant oriented bundles are mostly concerned with the engagement of the reader or the writer.

stance: giving an example

engagement: to engage the reader

the most frequent category in Computational Linguistics is the framing which indicates that most bundles are used in textual forms rather than topic specific bundles.

5.0 Appendix A: Complete list of Table 3.6

| | | |
|---|---|---|
| Research oriented | Location | the remainder of this section, while at the same time, as we will see in, that represents the meaning shown, in the right part of, shown in the right part, at the same time, at the end of, in this section we, in this chapter we, at the beginning of, is shown in figure, as shown in figure, as shown in table, as we will see, are shown in table, |
| | Procedure | this is not the case, the set of all possible, that it is possible to, for the purposes of this, we're going to |
| | | look at, it can be shown that, can also be used to, that can be used to is important to note that, can be used to, on the basis of, it is important to, as a function of, one of the most, as a result of, in such a way, it is necessary to, to take into account, as a sequence of, it is easy to, the ways in which, for the purposes of, with the help of, in the process of, can be used in, it is difficult to, the same way as, to be used in, |
| | Quantification | a very large number of, by the total number of, to reduce the number of, is the total number of, is the number of, the total number of, a large number of, a wide range of, of the number of, there are a number of, a little bit more, as the number of, the number of times, in a number of, to the number of, are a number of, a small number of, there are a number, by the number of, on the number of, of a number of, and the number of, a great deal of, |
| | Description | as a function of time, in the remainder of this, on the basis of, in the form of, to be able to, as a result of, the value of the, in the sense that, in the same way, can be viewed as, to the fact that, is said to be, be thought of as, be the set of, can be thought of, at the level of, is a set of, is referred to as, for the purposes of, the degree to which, for the purpose of, to be used in, |
| | topic | shown in the right part, that it is possible to, the value of the, to the science of, to take into account, in the study of, in the field of, can be applied to, can be used for, in the area of, in the course of, in the sense of, can also be used, for the purpose of, by the fact that, is a function of, is the same as, can be used as, an important role in, to the study of, this is the case |
| Text oriented | Transition | and on the other hand, on the one hand and, on the one hand |
| | Resultative | are more likely to be, can be thought of as, is going to be, can be found in, as a result of, the value of the, in such a way that, turn out to be, should be able to, as a sequence of, are going to be, turns out to be, it turns out that, more likely to be, is likely to be, are more likely to, can be interpreted as, |
| | Structuring | is due to the fact, we're going to talk about, on the other hand it, on the right hand side, in the rest of this, be the set of all, state of the art in, on the other hand is, it is well known that, as can be seen in, and you can see that, it is easy to see, in the |
| | | sense that they, the presence or absence of, is the set of all, and at the same time, on the other hand if, the state of the art, so on and so on, on the one hand and, and so on and so, with respect to the, due to the fact that, and you can see, of a set of, and so on and, for the sake of, in terms of their, of the fact that, from the fact that, on the use of, should be noted that, in a variety of, |
| | Framing | is often referred to as, there is no need to, the string of the form, in the same way that, may or may not be, in such a way as, less than or equal to, need to be able to, for the purposes of this, is one of the most, is the set of all, on the other hand if, it is not possible to, is important to note that, on the one hand and, as in the case of, in the case of, in the context of, it is possible to, the set of all, is the set of, if and only if, it is clear that, from the point of view, the point of view of, it is important to note, it should be noted that, the point of view, in the middle of, important to note that, and you can see, due to the fact¸ such a way that, the way in which, it is easy to, a wide variety of, the extent to which, it should be noted, the degree to which, in a way that, should be noted that, as a set of, |
| Participant oriented | Stance | such a way as to, this is an example of, one of the most important, can be seen as a, it is worth noting that, it can be shown that, this is going to be, it is also possible to, it is possible to, one of the most, there are a number of, as in the case, is an example of, that can be used, this is going to |
| | engagement | we're going to talk about, what i'm going to do, as can be seen from, it can be seen that, can be seen as a, we are able to, if we look at, if you want to, we can say that, going to look at, and you can see, as we will see, as can be seen, in this paper we, you can see that, can be seen as, so i'm going to, can be seen in, we would like to, so we're going to, and we're going to, |

6.0 Appendix B: List of Lexical bundles extracted from CL Academic Literature

| | | | |
|---|---|---|---|
| can be | to be | it is | in this |

for example
number of
such as
the same
this is
set of
is not
based on
of this
there is
we have
that is
of these
is to
there are
the following
going to
we can
which is
is that
has been
in which

use of
will be
may be
of language
used to
have been
the other
for each
of speech
one of
the number
the most
part of
does not
as well
that are
they are
the two
natural language
the first
of an
the second

are not
used in
would be
in order
kind of
be used
the word
and so
do not
and then
order to
need to
terms of
the system
in terms
well as
speech and
the language
which are
that we
rather than

should be
a set
a single
that it
we will
the case
the input
according to
types of
to use
as in
language and
a particular
is used
able to
the use
a number
if we
shown in
and in
as an
of english
each of

of words

and language

some of

to do

analysis of

the results

the set

for this

in figure

you can

must be

the data

so that

to have

the next

we are

associated with

and that

at least

have to

want to

is also

of all

the sentence

form of

the training

the speech

could be

what is

to make

a new

in proceedings

type of

for instance

fact that

the target

used for

the fact

of each

and to

in an

words in

if you

due to

in addition

in our

the probability

a given

the text

of their

or the

case of

in section

over the

a word

sequence of

the form

in some

note that

the model

that they

range of

this case

have the

approach to

second language

language learning

more than

in fact

that this

into a

the original

that there

the current

the best

a very

value of

that can

a more

are used

might be

computational linguistics

the main

training data
it has
of course
the problem
the previous
in english
in particular
is no
study of
in table
to this
respect to
not be
in fig
new york
conference on
the last
a and
the output
probability of

model of
level of
in general
the end
and their
this book
in other
look at
it can
related to
the way
example of
the study
in their
kinds of
but the
how to
the number of
in order to
in terms of
as well as
a set of

the use of
a number of
the fact that
based on the
in this case
the set of
there is a
can be used
with respect to
on the other
there is no
the case of
be used to
it is not
the other hand
and so on
in the case
we're going to

the end of
going to be
that it is
be able to
is based on
this is a
the probability of
that can be
in the following
in this section
in the same
it can be
i'm going to
a sequence of
is going to
the form of
the basis of
in addition to

the study of

that there is

the context of

we need to

the value of

shown in figure

the problem of

is used to

the process of

at the same

an example of

of the same

in other words

it is possible

the development of

the meaning of

for example in

the role of

a variety of

of the language

in the context

need to be

at the end

the results of

each of these

can also be

of the word

the notion of

and machine translation

on the basis

as shown in

the training data

the nature of

which can be

the size of

the same time

have to be

referred to as

of the two

it is important

of the form

the length of

as a result

can be seen

of the input

a lot of

a little bit

of the sentence

the performance of

can be found

in the form

more than one

the structure of

is possible to

the quality of

between the two

by means of

is to be

be found in

the output of

the relationship between

are used to

this section we

the beginning of

is important to

in the next

it is also

a function of

a list of

of the most

there are two

a series of

of the speech

the analysis of

the level of

in the previous

of the system

associated with the

has to be

the difference between

for example if

in this way

is shown in

analysis of the

this is not

the effect of

of the data

of the text

it would be

this kind of

in the second

of text and

the presence of

the amount of

the result of

needs to be

we want to

for a given

description of the

if you like

likely to be

it does not

that there are

you can see

i want to

page if you

point of view

book buy it

buy it page

is equal to

like this book

next page if

previous page page

this book buy

you like this

and it is

of a word

so this is

as part of

a range of

the ability to

there are many

the type of

in the first

and this is

it has been

the task of

one or more

the purpose of

it may be

the distribution of

in the sense

the same as

it should be

is one of

is that it

be used in

of language and

we do not

large number of

are based on

this means that

of the association

and speech and

but it is

in the sentence

the goal of

the rest of

to deal with

to be able

of this chapter

in the training

refer to the

the concept of

the most important

for each of

that we have

the sum of

the association for

of the target

in some cases

of a particular

with the same

on the web

shown in fig

shown in table

the choice of

be used for

of the utterance

this can be

on the other hand

in the case of

can be used to

on the basis of

at the same time

in the context of

it is possible to

at the end of

with respect to the

in this section we

it is important to

in the form of

book buy it page

if you like this

like this book buy

next page if you

page if you like

this book buy it

you like this book

is going to be

can be found in

to be able to

as a function of

one of the most

is the number of

the total number of

on the one hand

the set of all

as a result of

in this chapter we

at the beginning of

a large number of

the value of the

is shown in figure

in such a way

it is necessary to

is the set of

in the sense that

| | | | |
|---|---|---|---|
| as shown in figure | to the number of | the extent to which | it is important to note |
| in the same way | it is easy to | in the course of | the point of view of |
| to the science of | the ways in which | in the sense of | from the point of view |
| of text and speech | are a number of | of a set of | and so on and so |
| a wide range of | at the level of | can be seen as | can be thought of as |
| if and only if | in this paper we | for the purposes of | as in the case of |
| can be viewed as | is a set of | so i'm going to | on the one hand and |
| to the fact that | is an example of | a special case of | is important to note that |
| it is clear that | a small number of | are shown in table | so on and so on |
| of the number of | a wide variety of | with the help of | the state of the art |
| evaluation of text and | can be applied to | and so on and | it is not possible to |
| it turns out that | can be used for | it should be noted | on the other hand if |
| such a way that | there are a number | in the process of | it is also possible to |
| the number of times | in the area of | in such a way that | this is going to be |
| in a number of | that can be used | there are a number of | that can be used to |
| the way in which | you can see that | it should be noted that | and at the same time |
| as can be seen | is referred to as | due to the fact that | in the same way as |

can also be used to

it can be shown that

is the set of all

it is worth noting that

the presence or absence of

we're going to look at

is one of the most

that it is possible to

the set of all possible

this is not the case

in the remainder of this

can be seen as a

in the right part of

shown in the right part

one of the most important

this is an example of

is the total number of

it can be seen that

that represents the meaning shown

are more likely to be

as can be seen from

for the purposes of this

need to be able to

as we will see in

in the sense that they

it is easy to see

less than or equal to

the picture this is a

and you can see that

as can be seen in

in such a way as

may or may not be

of the picture this is

part of the picture this

right part of the picture

in the same way that

it is well known that

on the other hand is

state of the art in

be the set of all

in the rest of this

on the right hand side

the string of the form

there is no need to

to reduce the number of

on the other hand it

such a way as to

what i'm going to do

while at the same time

and on the other hand

as a function of time

by the total number of

we're going to talk about

a very large number of

the remainder of this section

is due to the fact

is often referred to as